\documentclass[letterpaper]{article}
\usepackage[noadjust]{cite}
\usepackage{aaai20}
\usepackage{times}
\usepackage{helvet}
\usepackage{courier}
\usepackage[hyphens]{url} 
\usepackage{graphicx}
\title{Generalized Hidden Parameter MDPs:\\Transferable Model-based RL in a Handful of Trials}
\author{
  Christian F. Perez, Felipe Petroski Such, Theofanis Karaletsos\\
  Uber AI Labs\\
  San Francisco, CA 94105\\
  \texttt{\{cfp,felipe.such,theofanis\}@uber.com}\\
}
\usepackage{tikz}
\usetikzlibrary{bayesnet}
\usepackage{nicefrac}       % compact symbols for 1/2, etc.
\usepackage{microtype}      % microtypography
\usepackage{amsfonts}
\usepackage{amssymb}
\usepackage{amsmath}
\usepackage{algorithm}
\usepackage{algpseudocode}
\usepackage{bm}
\usepackage{xcolor}
\usepackage[title]{appendix}
\usepackage[utf8]{inputenc} % allow utf-8 input
\usepackage[T1]{fontenc}    % use 8-bit T1
\usepackage{comment}

\newcommand{\reals}{\mathbb{R}}
\newcommand{\s}{\mathbf s_t}
\newcommand{\ns}{\mathbf s_{t+1}}
\newcommand{\action}{\mathbf a_t}
\newcommand{\env}{\mathbf z}
\newcommand{\latR}{\mathbf z_r}
\newcommand{\latE}{\mathbf z_d}
\newcommand{\latA}{\mathbf z_a}
\newcommand{\lat}{\mathbf z}

\newcommand{\return}{\mathbf R}
\newcommand{\dparams}{{\theta}}
\newcommand{\rparams}{{\omega}}
\newcommand{\defines}{\;\dot{=}\;}
\newcommand{\frew}{\mathcal R}
\newcommand{\rtp}{\mathbf r_{t+1}}
\newcommand{\episode}{T}

\newcommand{\transition}{\mathcal T}

\definecolor[named]{green}{rgb}{0.1,0.5,0.1}

\newcommand{\expname}[1]{\mbox{\texttt{\textls[-60]{#1}}}}
\begin{document}

\maketitle

\begin{abstract}
There is broad interest in creating RL agents that can solve many (related) tasks and adapt to new tasks and environments after initial training. Model-based RL leverages learned surrogate models that describe dynamics and rewards of individual tasks, such that planning in a good surrogate can lead to good control of the true system.  Rather than solving each task individually from scratch, hierarchical models can exploit the fact that tasks are often related by (unobserved) causal factors of variation in order to achieve efficient generalization, as in learning how the mass of an item affects the force required to lift it can generalize to previously unobserved masses. We propose Generalized Hidden Parameter MDPs (GHP-MDPs) that describe a family of MDPs where both dynamics and reward can change as a function of hidden parameters that vary across tasks. The GHP-MDP augments model-based RL with latent variables that capture these hidden parameters, facilitating transfer across tasks. We also explore a variant of the model that incorporates explicit latent structure mirroring the causal factors of variation across tasks (for instance: agent properties, environmental factors, and goals). We experimentally demonstrate state-of-the-art performance and sample-efficiency on a new challenging MuJoCo task using reward and dynamics latent spaces, while beating a previous state-of-the-art baseline with $>10\times$ less data. Using test-time inference of the latent variables, our approach generalizes in a single episode to novel combinations of dynamics and reward, and to novel rewards.
\end{abstract}

\section{Introduction}
\label{introduction}

In our pursuit of better learning algorithms, we seek those that can learn quickly across tasks that it encounters during training (called positive \emph{transfer} when there is helpful synergy), and generalize to novel it encounters at test-time. Consider an illustrative problem of an agent with some pattern of broken actuators (agent variation) acting in an environment with changing surface conditions due to weather (dynamics variation), tasked with achieving one of many possible goals (reward variation). We would like a learning algorithm that (1) pools information across observed tasks to learn faster (positive transfer), and generalizes from observed combinations of agent, dynamics, and reward variations to (2) other unseen combinations (\emph{weak} generalization) and (3) novel variations (\emph{strong} generalization) without learning a new policy entirely from scratch \cite{synpo}.

To tackle this problem, we introduce Generalized Hidden Parameter MDPs (GHP-MDP) to describe families of MDPs in which dynamics and reward can change as a function of hidden parameters (Section~\ref{sec:GHP-MDP}). This model introduces (multiple) latent variables that capture the factors of variation implicitly represented by tasks at training time. At test time, we infer the MDP by inferring the latent variables that form a latent embedding space of the hidden parameters.
This extends and unifies two lines of work: we augment transferable models of MDPs like \cite{doshi2016hidden} with structure on reward and dynamics, and combine it with powerful approaches for learning probabilistic models \cite{balaji2017} to solve challenging RL tasks.

We propose two variants of latent variable models: one with a shared latent variable to capture all variations in reward and dynamics, and a structured model where latent variables may factorize causally. Our intention is to afford GHP-MDPs the use of prior knowledge and inductive biases that could improve sample efficiency, transfer, and generalization.

In our experiments, agents are trained on a small subset of possible tasks---all related as instances from the same GHP-MDP---and then generalize to novel tasks from the same family via inference. We show that this method improves on the state-of-the-art sample efficiency for complex tasks while matching performance of model-free meta-RL approaches (Section~\ref{sec:experiments}) \cite{finn2017,PEARL}. Notably, our approach also succeeds with a fairly small number of training tasks, requiring only a dozen in these experiments.

\section{Model-based RL}

We first consider a reinforcement learning (RL) problem described by a Markov decision process (MDP) comprising a state space, action space, transition function, reward function, and initial state distribution: $\{\mathcal{S}, \mathcal{A}, \transition, \frew, \rho_0 \}$ \cite{Bellman1957}. We define a "task" (or "environment") $\tau$ to be a MDP from a set of MDPs that share $\mathcal{S}$ and $\mathcal{A}$ but differ in one or more of $\{ \transition, \frew, \rho_0 \}$.

In model-based RL, the agent uses a model of the transition dynamics $\transition: \mathcal{S}\times \mathcal{A} \rightarrow \mathcal{S}$ in order to maximize rewards over some task-dependent time horizon $H$. For a  (stochastic) policy $\pi_\theta$ parameterized by $\theta$, the goal is to find an optimal policy $\pi^*$ that maximizes the expected reward,
\begin{equation*}
\begin{split}
\pi^*(\mathbf{a}|\mathbf{s}) = &\mathrm{argmax}_\theta\,\mathbb{E}_{\action \sim \pi_\theta(\cdot|\s)} \sum_{t'=t}^{t+H-1} \frew(\mathbf{s}_{t'}, \mathbf{a}_{t'}) \\
\quad &\mathrm{s.t.}\, \ns \sim \transition(\s,\action),
\end{split}
\end{equation*}
where $\transition$ acts as a probability distribution over next states in a stochastic environment.

While it is common to assume a known reward function $\frew$ and even transition function $\transition$, one can simultaneously learn an approximate model of both the dynamics and reward,
\begin{align*}
\transition \approx \tilde{\transition} &\defines p_{\dparams}(\ns|\s,\action)\\
\frew \approx \tilde{\frew} &\defines p_{\rparams}(\rtp|\s,\action, \ns)
\end{align*}
with parameters $\dparams$ and $\rparams$ using data collected from the environment $\mathcal D = \{(\mathbf{s}_t^{(n)},\mathbf{a}_t^{(n)},\mathbf{s}_{t+1}^{(n)},r^{(n)}_{t+1})\}_{n=1}^N$. In this work, we do not learn a parametric policy $\pi_\theta$, but instead use model predictive control to perform planning trajectories sampled from the learned models (see Section~\ref{sec:control}.)

The RL problem is then decomposed into two parts: learning models from (limited) observations, and (approximate) optimal control given those models. 
% We consider each of these components from a probabilistic perspective in Section~\ref{sec:rl-as-inference}. 
By iterating between model learning and control, the agent uses the improved model to improve control and vice versa. This basic approach is effective for a single environment, but is not designed to learn across multiple related environments. This is a key limitation our approach overcomes, described in Section~\ref{sec:GHP-MDP}.

Another shortcoming of this approach is the tendency for expressive models (e.g., neural networks) to overfit to observed samples and produce overconfident and erroneous predictions (also called "model bias" ~\cite{modelbias}). The result is a sub-optimal policy, worse sample efficiency, or both. This problem is exacerbated for transfer learning scenarios, in which an agent that overfits to training tasks fails to generalize to novel tasks at test time. Bayesian learning for neural networks can properly account for model uncertainty given limited data, but can be difficult to scale to high-dimensional states $\mathcal{S}$ and actions $\mathcal{A}$ \cite{Deisenroth2011a} and is burdened by the hardness of representation for the posterior over weights. As a tractable alternative, we extend \emph{Deep Ensembles} of neural networks~\cite{balaji2017}, which perform well on isolated tasks \cite{Chua2018}, to learn transferable agents without changing the underlying model.

\subsection{Learning probabilistic models}
\label{sec:models_ensembles}

In order to perform model-based control, an agent requires knowledge of the dynamics $p(\ns | \s, \action)$ and reward $p(\rtp | \s, \action, \ns)$. When these underlying mechanisms are unknown, one can resort to learning parameterized models $p_\dparams(\ns|\s,\action)$ and $p_\rparams(\rtp |\s,\action,\ns)$. Because environments can be stochastic, we use a generative model of dynamics and reward.
Because these are continuous quantities, each can be modeled with a Gaussian likelihood. The dynamics, for example, can be parameterized by mean $\bm\mu_\dparams$ and diagonal covariance $\bm\Sigma_\dparams$ produced by a neural network with parameters $\dparams$ (and similarly for the reward model using parameters $\rparams$),
\begin{equation}
\begin{alignedat}{2}
p_\dparams(\ns|\s,\action) &= \mathcal{N}(& &\bm{\mu}_\dparams(\s, \action), \bm\Sigma_\dparams(\s, \action)) \\
p_{\rparams}(\rtp|\s,\action,\ns) &= \mathcal{N}(& &\bm{\mu}_\rparams(\s,\action,\ns) ,\\
& \phantom{{}={}} & &\bm{\Sigma}_\rparams(\s,\action,\ns)) \,.
\end{alignedat}
\label{eq:likelihood}
\end{equation}
From these building blocks, we construct a joint probability distribution over trajectories and jointly optimize model parameters $\{\dparams, \rparams\}$ given data $\mathcal D$. (See Section~\ref{sec:training-inference} for the learning objective involving $\bm \mu_\theta$ and $\bm\Sigma_\theta$.)

Also, as commonly done elsewhere, the neural network prediction target is actually the change in the states ${\Delta_s = \ns - \s}$ given the state and action: $p_\dparams(\Delta_s | \s, \action;\theta)$.

\subsection{Ensembles of networks}

In order to be robust to model misspecification and handle the small data setting, one can model uncertainty about parameters $\dparams$ and $\rparams$ and marginalize over their posterior after observing dataset $\mathcal{D}$ to obtain the predictive distributions
{\small
\begin{equation}
\begin{split}
p(\ns|\s,\action, \mathcal{D}) &= \int p_\dparams(\ns|\s,\action) p(\dparams|\mathcal{D})  d\dparams\\
p(\rtp|\s,\action,\ns, \mathcal{D}) &= \int p_\rparams(\rtp|\s,\action,\ns) p(\rparams|\mathcal{D})  d\rparams \,.
\label{eq:bnn}
\raisetag{33pt}
\end{split}
\end{equation}
}
Learning these models can be posed as inference of model parameters given observed data, e.g. using a Bayesian Neural Network (BNN) which entails inferring the posteriors $p(\dparams|\mathcal{D})$ and $p(\rparams|\mathcal{D})$. 
As is usually the case in inference for such models, computing the exact posterior is intractable. A practical way to approximate the predictive distribution of the network is by capturing uncertainty through frequentist ensembles of models, in which each ensemble member is trained on a shuffle of the training data \cite{balaji2017}. For an ensemble with $M$ members and the collection of all network parameters $\Theta \defines \{\theta_1,...,\theta_{M}\}$, we define a model of the next state predictive distribution as a mixture model as follows:
\begin{equation}
\begin{split}
    p(\ns|\s,\action; \Theta) &= \frac{1}{M}\sum \limits_{\theta \in \Theta} p_\theta(\ns|\s,\action) \\
    &\approx p(\ns|\s,\action) \,.
\end{split}
\label{eq:ensemble}
\end{equation}
The reward model follows similarly,
\begin{equation}
\begin{split}
    p(\rtp|\s,\action,\ns;\Omega) &= \frac{1}{M} \sum\limits_{\rparams \in \Omega} p_\rparams(\rtp|\s,\action,\ns) \\
    &\approx p(\rtp|\s,\action,\ns) \,,
\raisetag{12pt}
\end{split}
\end{equation}
also including its dependence on $\ns$, whose values are observed from training data, but at test-time are the result of predictions from they dynamics model of \eqref{eq:ensemble}.
\begin{figure}
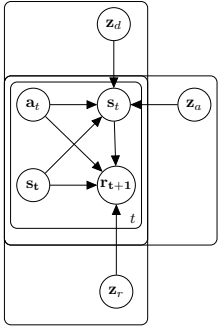

\centering
\resizebox{0.35\linewidth}{!}{
\tikz{
    % Nodes
    \node[latent] (s)   {$\bf{s}_t$};
    \node[latent, above=of s]   (a)   {$\action$}; 
    \node[latent, right=of a]   (ns) {$\s$};
    \node[latent, above=of ns] (z) {$\latE$};
    \node[latent, right=of ns] (za) {$\latA$};
    \node[latent, right=of s] (r) {$\bf{r_{t+1}}$};
    \node[latent, below=1.5 of r] (zr) {$\latR$};
    
    \edge {s,a,z,za} {ns};
    \edge {s,a,ns,zr} {r};
   
    \plate {mdp} {(s)(ns)(r)(a)} {$t$};
    % \tikzset{plate caption/.append style={below right=0pt and 0pt of #1.north west}
    \plate {env} {(z)(mdp)} {};
    \plate {agent} {(za)(mdp)} {};
    \plate {reward} {(zr)(mdp)} {};
}
}
\caption{Probabilistic graphical model of MDP with structured latent variables for environment, agent, and reward function variations.}
\label{fig:model2}
\end{figure}

\section{Modeling with hidden parameters}

Our goal is to learn a model of the system we would like to control and then plan on that model in order to achieve high reward on the actual system. For sufficiently complex systems and finite training data, we expect that the model can only approximate the real system. Furthermore, the real system may differ in significant ways from the system our models were trained on, as when a robot actuator force degrades over time, unless the conditions were deliberately included in training. However, it is unreasonable to train a model across all possible conditions an agent may encounter.
Instead, we propose a model that learns to account for the causal factors of variation observed across tasks at training time, and then infer at test time the model that best describe the system. We hypothesize that explicitly incorporating these factors will facilitate generalization to novel variations at test time.

A Hidden Parameter MDP (HiP-MDP), first formalized in \cite{doshi2016hidden}, describes a family of MDPs in which the transition dynamics are parameterized by hidden parameters $\eta \in \reals^n$, here expressed as $\transition_\eta$ or $\transition(\cdot\,;\eta)$. In dynamical systems, for example, parameters can be physical quantities like gravity, friction of a surface, or the strength of a robot actuator. Their effects are "felt" but not directly observed; $\eta$ is not part of the observation space.
Prior work using meta-learning for adaptive dynamics problems proposes agents that learn across \emph{distinct} tasks $\tau_i$ where the transition dynamics varies according to some problem-specific distribution $\transition_i \sim p(\transition)$. For example, by changing the terrain, agent ability, or observations for tasks during training, agents can learn to adjust to novel yet similar conditions at test time \cite{Clavera2018,Fu2016}.
In contrast, we treat each task as an instance of the same HiP-MDP with different hidden parameters $\eta \sim p(\eta)$ affecting the transition function $\transition_\eta$. Previous work showed that for simple, low-dimensional systems, agents can infer an effective representation of the hidden parameters, and generalize the dynamics to novel parameter settings \cite{killian2017robust,saemundsson2018meta}. Yet neither the HiP-MDP nor adaptive dynamics methods account for different task rewards.

Consider a multi-task setting, in which an agent learns across tasks $\tau_i$ where only the reward function $\frew_i$ varies, for example, performing tasks that require navigation to a goal position, or movement in a certain direction or target velocity \cite{finn2017}. In our formulation, all of these tasks come from a parameterized MDP in which the reward function $\frew_\eta$ or $\frew(\cdot; \eta)$ depends on hidden parameters $\eta$ that determine the goal/reward structure.

\subsection{Generalized Hidden Parameter MDPs}
\label{sec:GHP-MDP}

We denote a set of tasks/MDPs with transition dynamics $\transition_\eta$ and rewards $\frew_\eta$ that are fully described by hidden parameters $\eta$ as a {\it Generalized Hidden Parameter MDP} ({\it GHP-MDP}). 
A GHP-MDP includes settings in which tasks can exhibit multiple factors of variation. For example, consider a robotic arm with both an unknown goal position $g$ and delivery payload $m$. This problem can be modeled as drawing tasks from a distribution $\eta_g, \eta_m \sim p(\eta)$ with effects on both the transition $\transition_{\eta_m}$ and reward function $\frew_{\eta_g}$.
Additional factors of variation can be modeled with additional parameters, for example, by changing the size of the payload $\eta_l$. Note that we generalize $\eta$ to describe more than just physical constants. All of these hidden parameters are treated as latent variables $\{\lat_i \in \reals^{d_i} : i=1,\ldots,c \}$, and we express the GHP-MDP as a latent variable model.

We pose jointly learning the two surrogate models of \eqref{eq:likelihood} and latent embeddings $\lat_i$ via the maximization of a variational lower bound over data collected from a small set of training tasks (see Section~\ref{sec:training-inference} for the inference objective.) At test-time, only the parameters $\phi$ for the approximate posterior $p_\phi(\lat_i|\mathcal D)$ of the latent variables are learned via inference. Note that the latent variables $\lat_i$ are an \emph{embedding} of the true parameters $\eta$, and in general, are not equal to $\eta$, nor even have the same dimensions (i.e., $d_i \neq n$).

In Section~\ref{sec:latent-model}, we describe the simplest probabilistic model of a GHP-MDP that uses a single continuous latent variable $\lat \in \reals^D$ to model hidden parameters of both the dynamics and reward. Because a single $\lat$ jointly models all unobserved parameters, we call this the \emph{joint latent variable (joint LV)} model. In Section~\ref{sec:plated-model}, we extend the model to multiple latent variables $\{\lat_d,\lat_a,\lat_r \} \in \reals^D$ (shown graphically in Figure~\ref{fig:model2}), one for each aspect of the task that is known to vary in the training environments. In other words, we encode our prior knowledge about the (plated) structure of the tasks into the structure of the model; hence, we refer to this as the \emph{structured latent variable (structured LV)} model. In this paper, we assume that latent variables are specified \emph{a prior} to be either shared or distinct. We leave learning how to disentangle these factors to future work.

\subsection{Joint latent variable model}
\label{sec:latent-model}

For any GHP-MDP, we can model the dynamics and reward hidden parameters jointly with a single latent variable $\env \in \reals^D$. The model for episode return $\return=\sum \rtp$ for a trajectory decomposed into partial rewards $\rtp$ is
\begin{equation}
p(\return|\mathbf{s}_{0:T}, \mathbf{a}_{0:T-1}, \env) = \prod \limits_{t=0}^{T-1} p_\rparams(\rtp|\s, \action, \ns, \env)\,,
\end{equation}
where $T$ is the episode length. The resulting joint model mapped over trajectories $p( \mathbf{s}_{0:T}, \mathbf{a}_{0:T-1}, \return, \lat)$ is
\begin{equation}
\begin{split}
p(\env) p(\mathbf s_0) \prod \limits_{t=0}^{T-1}\big[& p(\rtp|\s, \action, \ns, \env) \\ 
& p(\ns|\s,\action, \env) p(\action|\s,\env)\big]
\label{eq:dynamics}
\end{split}
\end{equation}
The key feature of this model is the same latent variable $\lat$ conditions both the dynamics and the reward distributions. The priors for auxiliary latent variable are set to simple normal distributions, $p(\env) = \mathcal{N}(\bf{0},\bf{I})$, and initial state distribution $p(\mathbf s_0)$ to the environment simulator. Again we note that $p(\action|\s,\env) = \pi^{*}(\action|\s,\env)$ via planning (Section~\ref{sec:control}).

Common meta-RL tasks that vary only reward/task, e.g. by varying velocities or goal directions \cite{finn2017}, can be tackled with a simplified joint LV model that conditions only the reward model. Combined with inference at test-time, this approach is an alternative to meta-learning via MAML or RNNs \cite{Duan2017}. Our experiments (Section~\ref{sec:experiments}) demonstrate the full joint LV model (conditioning both dynamics and reward models) can solve one such reward-variation task.

\subsection{Structured latent variable model}
\label{sec:plated-model}

In the previous section, we introduced a global latent variable $\lat$ that is fed into both the dynamics and reward model.
Here, we extend this idea and introduce multiple {\it plated variables} which constitute the structured latent space of the GHP-MDP. Separate latent spaces for dynamics and reward are intuitive because agents may pursue different goals across environments with different dynamics.

Consider a structured model with two latent variables $\latE \in \reals^{D_1}$ and $\latR \in \reals^{D_2}$ to separately model hidden parameters in the dynamics $\transition(\cdot\,;{\latE})$ and reward $\frew(\cdot\,;{\latR})$.
The joint model $p(\mathbf{s}_{0:T+1}, \mathbf{a}_{0:T}, \return, \latE, \latR)$, including the action distribution implied by control, is
\begin{align}
\begin{split}
p(\mathbf{z}_d) p(\mathbf{z}_r)
p(\mathbf s_0) \prod \limits_{t=0}^{\episode - 1} \big[& p(\rtp|\s,\action,\ns, \latR) \\
&p(\ns|\s,\action, \latE) p(\action|\s,\latR,\latE) \big]\,.
\raisetag{38pt}
\end{split}
\end{align}
This structure facilitates solving tasks where both of these aspects can vary independently. Typically, only one or the other is varied in meta-RL tasks for continuous control \cite{Clavera2018,finn2017,PEARL}, and so we introduce new tasks in Section~\ref{sec:experiments} to test the weak and strong generalization ability of this modeling choice.

More generally we may have $c$ arbitrary plated contexts, such as agent, dynamics, reward variation, etc. Then for the set of latent variables $\{\mathbf{z}_1 , \ldots, \mathbf{z}_c \}$, each explains a different factor of variation in the system, implying $p(\lat) = \prod p(\mathbf{z}_c)$.
This allows the model to have separate degrees of freedom in latent space for distinct effects. Note that the use of plated variables implies that tasks will have known factors of variation (but unknown values and effects) at training time only. In practice, this is case when training on a simulator.

By factorizing the latent space to mirror the causal structure of the task, the structured LV model can also more efficiently express the full combinatorial space of variations.
For example, with $c=3$ factors of variation and 10 variations for each $\eta_i$ for $i \in \{1, 2, 3\}$, the latent space must generalize to $10 \times 10 \times 10 = 10^3$ combinations. Learning a global latent space would require data from some non-trivial fraction of this total. In contrast, a structured space can generalize from $10 + 10 + 10 = 30$. We examine this generalization ability experimentally in Section~\ref{sec:strong-generalization}.

\begin{algorithm*}[htb]
\caption{Learning and control with MPC and Latent Variable Models}
\label{al:mpc}
\begin{algorithmic}[1]
\State Initialize data $\mathcal{D}$ with random policy
\For{Episode m = 1 to M}
    \State Sample an environment indexed by $k$
    \State If learning, train a dynamics model $p_\dparams$ and reward model $p_\rparams$ with
    $\mathcal{D}$ using \eqref{eq:loss-joint} or \eqref{eq:loss-plated}
    \label{al:train}
    \State Initialize starting state $\mathbf s_0$ and episode history $\mathcal{D}_k = \varnothing$
    \For{Time t = 0 to TaskHorizon}
        \Comment{MPC loop}
        \For{Iteration i = 0 to MaxIter}
            \Comment{CEM loop}
            \State Sample actions $\mathbf{a}_{t:t+h} \sim \mathrm{CEM}(\cdot)$
            \label{al:cem-sample}
            \State Sample latent $\lat^{(p)} \sim q_\phi(\env)$ for each state particle $s_p$
            \State Propagate next state  $\mathbf{s}^{(p)}_{t+1} \sim p_{\dparams}(\cdot | \mathbf{s}_{t}^{(p)}, \action, \lat^{(p)})$ using TS-$\infty$
            \Comment{See \cite{Chua2018}}
            \label{al:state-sample}
            \State Sample reward $\rtp^{(p)} \sim p_\rparams(\cdot|\s^{(p)},\action,\ns^{(p)},\lat^{(p)})$ for each particle trajectory
            \State Evaluate expected return $G_t = \sum\limits_{\tau=t}^{t+h} \nicefrac{1}{P}\sum\limits_{p=1}^P \mathbf{r}_{\tau+1}^{(p)}$
            \If {$\operatorname{any}(\operatorname{early\_termination}(\s^{(p)}))$}
                \State $G_t \gets \mathrm{done\_penalty}$
                \Comment{Hyperparameter for early termination}
            \EndIf
            \State Update $\mathrm{CEM}(\cdot)$ distribution with highest reward trajectories
            \EndFor
        \State Execute first action $\action$ determined by final $\mathrm{CEM}(\cdot)$ distribution
        \State Record outcome $\mathcal{D}_t \gets  \{(\s,\action)), (\ns,\rtp)\} $
        \State Record outcome $\mathcal{D}_k \gets \mathcal{D}_k \cup \mathcal{D}_t $
        \State If test-time, update approximate posterior $q_\phi(\env|\mathcal{D}_t)$ using \eqref{eq:loss-joint}
    \EndFor
    \State Update data $\mathcal{D} \gets \mathcal{D} \cup \mathcal{D}_k$
\EndFor
\end{algorithmic}
\end{algorithm*}

\subsection{Training and inference}
\label{sec:training-inference}
{Each step/episode of training consists of two phases: collect an episode of trajectories $\mathcal D_k$ for each task via planning (Algorithm~\ref{al:mpc}), and infer model parameters and latent variables using \emph{all} collected data via SGD. The goal of the inference (learning) step is to maximize the marginal likelihood of observed transitions with respect to $\theta$ and $\phi$. For the \emph{joint latent variable model}, the intractable distribution $p(\env|\mathcal{D})$ is approximated with $q_\phi(\env)$ parameterized by a diagonal Gaussian where ${\phi}=\{\mu, \Sigma\}$.} We then maximize the evidence lower bound ({\bf ELBO}) to the marginal log-likelihood:
{\small
\begin{multline}
    \log p(\mathcal D) = \sum_{t=0}^T \big[\log p(\ns|\s,\action) + \log p(\rtp|\s,\action,\ns) \big] \\
    \geq \mathbb{E}_{\env \sim q_\phi(\env)} \left[ \sum_{t=0}^{T} \left( \log p_\dparams(\ns|\s,\action,\env) + \right. \right. \\
    \left. \vphantom{\sum_{t=0}^{T}} \left. \log p_\rparams(\rtp|\s,\action,\ns,\env) \right) \right] - \mathrm{KL}\big(q_\phi(\env) || p(\env)\big).
\label{eq:objective}
\end{multline}
}
For simplicity, we choose the prior $p(\env)$ and variational distribution $q_\phi(\env)$ to be Gaussian with diagonal covariance. We can use this criterion during the training phase to jointly update network parameters $\Theta$ and variational parameters $\phi$ capturing beliefs about latent variables.

In practice, we use stochastic variational inference \cite{ranganath2013black,kingma2014stochastic} and subsample in order to perform inference and learning via gradient descent, yielding the loss function:
\begin{equation}
\begin{split}
\mathcal{L}(\dparams, \rparams, \phi) = \\
-\frac{1}{M} \sum \limits_{m=1}^{M} \sum_{t=0}^{T} \big[ & \log p_\dparams(\ns|\s,\action,\env^{(m)}) \\
&+ \log p_\rparams(\rtp|\s,\action,\ns,\env^{(m)}) \big] \\
+ \mathrm{KL}\big(q_\phi(\env) || p(\env)\big) 
\label{eq:loss-joint}
\raisetag{10pt}
\end{split}
\end{equation}
with $\env^{(m)} \sim q_\phi(\env)$ and number of samples $M=2$. {Recall that both models are ensembles and each network in the ensemble is optimized independently, but the variational distribution is shared according the relationship between tasks. During training, we minimize $\mathcal{L}(\dparams, \rparams, \phi)$, and at test time, reset $q_\phi$ to the prior and minimize with respect to $\phi$ only.}

For structured latent variable models with plated contexts, \eqref{eq:loss-joint} can be extended to multiple latent variables. For a graphical model (Fig.~\ref{fig:model2}) with three factors of variation---environment, agent, and reward---and variational parameters $\Phi \defines \{\phi_d,\phi_a,\phi_r\}$ for each, the loss function becomes
{\small
\begin{equation}
\begin{split}
 \mathcal{L}(\dparams, \rparams, \phi) = \\
 -\frac{1}{M} \sum \limits_{m=1}^{M} \sum_{t=0}^{T} \big[ & \log p_\dparams(\ns|\s,\action,\latE^{(m)},\latA^{(m)}) \\
  &+ \log p_\rparams(\rtp|\s,\action,\ns,\latR^{(m)}) \big] \\
 &+\mathrm{KL}(q_{\phi_d}(\latE) || p(\latE)) \\
 &+\mathrm{KL}( q_{\phi_a}(\latA) || p(\latA) )\\
 &+\mathrm{KL}( q_{\phi_r}(\latR) || p(\latR) )\,.
\label{eq:loss-plated}
\end{split}
\end{equation}
}

\subsection{Control with Latent Variable Models}
\label{sec:control}

Given a learned dynamics model, agents can plan into the future by recursively predicting future states $\mathbf s_{t+1}, ..., \mathbf s_{t+h}$ induced by proposed action sequences $\mathbf a_t, \mathbf a_{t+1}, ..., \mathbf a_{t+h}$ such that $\ns \sim \tilde{\transition}(\s,\action)$. If actions are conditioned on the previous state to describe a policy $\pi(\action|\s)$, then planning becomes learning a policy $\pi^{*}$ to maximize expected reward over the predicted state-action sequence. A limitation of this approach is that modeling errors are compounded at each time step, resulting in sub-optimal policies when the learning procedure overfits to the imperfect dynamics model. Alternatively, we use \emph{model predictive control (MPC)} to find the action trajectory $\mathbf a_{t:t+H}$ that optimizes $\sum_t^{t+H-1} \mathbb{E}_{q_\phi(\lat)}\mathbb{E}_{p(\mathbf{s_{t}},\mathbf{a}_{t})}[p(\rtp|\s, \action,\ns, \env)]$ at run-time \cite{Camacho}, {using $\ns$ predicted from the learned model (Algorithm~\ref{al:mpc}, line \ref{al:state-sample})}. At each time step, the MPC controller plans into the future, finding a good trajectory over the planning horizon $H$ but applying only the first action from the plan, and re-plans again at the next step. Because of this, MPC is better able to tolerate model bias and unexpected perturbations.

Algorithm~\ref{al:mpc} includes a control procedure that uses the cross-entropy method (CEM) as the optimizer for an MPC controller \cite{cem_tutorial}. On each iteration, CEM samples 512 proposed action sequences $\mathbf a_{t:t+H-1}$ from $H$ independent multivariate normal distributions $\mathcal N (\action | \bm\mu_t,\bm\Sigma_t)$, one for each time step in the planning horizon (line~\ref{al:cem-sample}), and calculates the expected reward for each sequence. The top 10\% performing of these are used to update the proposal distribution mean and covariance. However, evaluating the expected reward exactly is intractable, so we use a particle-based approach called trajectory sampling (TS) from \cite{Chua2018} to propagate the approximate next state distributions. 
We adapt the TS+CEM algorithm to incorporate beliefs about the MDP given data observed so far:
Each state particle $\s^{(p)}$ uses a sample of each latent variable $\lat^{(p)} \sim q_{\phi}(\lat)$ so that planning can account for their effect on the dynamics and reward models.

At test time, we skip line~\ref{al:train} to keep the neural networks fixed. The algorithm iterates between acting in the environment at step $t$ and inferring $p(\env | \mathcal{D}_{t})$ in order to align the dynamics and reward models with the current system as new information is collected.
In order to plan when episodes can terminate early due to constraints set by the environment (e.g., when MuJoCo Ant or Walker2d falls over), we set cumulative rewards for particle trajectories that violate those constraints to a fixed constant. This hyperparameter is set to $0$ during training to allow exploration, and $-100$ at test time for more conservative planning.

\section{Related Work}

Transfer learning and learning transferable agents has a long history in reinforcement learning; see \cite{taylor2009transfer,lazaric2012transfer} for a survey.  Recent prior work on latent variable models of MDPs focused on models of dynamics with small discrete action spaces \cite{doshi2016hidden,killian2017robust,yao2018}. We extend Hidden Parameter MDPs introduced in~\cite{doshi2016hidden} to the more general case including reward modeling and accounting for potentially multiple factors of variation. Latent dynamics models for hard continuous control tasks were used in \cite{WORKSHOP}, and using Gaussian Processes in~\cite{saemundsson2018meta}, albeit under significantly less challenging experimental conditions. Another notable use is latent skill embeddings in robotics for adaptation to different goals \cite{hausman2018}. In contrast, \emph{DeepMDP} models a single task/MDP entirely in a latent space \cite{deepmdp}, and can be merged to form a Deep-GHP-MDP. Similarly, \cite{hafner2019} demonstrates latent space planning from pixel observations, and has some multi-task ability when the tasks appear different. \cite{zhang2018} also learns and plans in a latent space with learned dynamics and reward models, and explores transfer using prior learned encoders on slightly perturbed tasks. The problem of learning and generalization across combinations of reward and dynamics in discrete action spaces on visual domains was also tackled in \cite{synpo} using factorized policies and complementary embeddings of reward and dynamics.

Meta-learning, or learning to learn, is one solution that enables RL agents to learn quickly across different or non-stationary tasks \cite{Schmidhuber1987,Ravi2017,AlShedivat2017ContinuousAV,finn2017}. Recently, meta-learning has been used to adapt a dynamics model \cite{Clavera2018} for model-based control to changing environments (but does not model reward or solve multiple tasks), or to learn a policy \cite{Rothfuss2018,PEARL} that adapts by adjusting the model or policy in response to recent experience. Extensions can continuously learn new tasks online \cite{nagabandi2018deep}. However, model-free meta-RL methods can require millions of samples and dozens of training tasks. One can also simply adapt a neural network online at test-time via SGD without MAML for one/few-shot learning \cite{Fu2016}.
Learning across tasks with common dynamics has been approached with meta-reinforcement learning \cite{finn2017} or using successor features \cite{successor}.

\section{Experiments}
\label{sec:experiments}

{
The GHP-MDP conceptually unifies various RL settings, including multi-task RL, meta-RL, transfer learning, and test-time adaptation. In these experiments, we wish to 1) demonstrate inference on a GHP-MDP with multiple factors of variation, 2) compare the sample efficiency and performance of our model-based implementation of the GHP-MDP on hard continuous control tasks with these flavors, and 3) explore its ability to generalize to novel tasks at test time.

\subsection{Didactic example with multiple factors}
\label{sec:toy}

\begin{figure}[ht]
    \centering
    \includegraphics[width=.95\columnwidth]{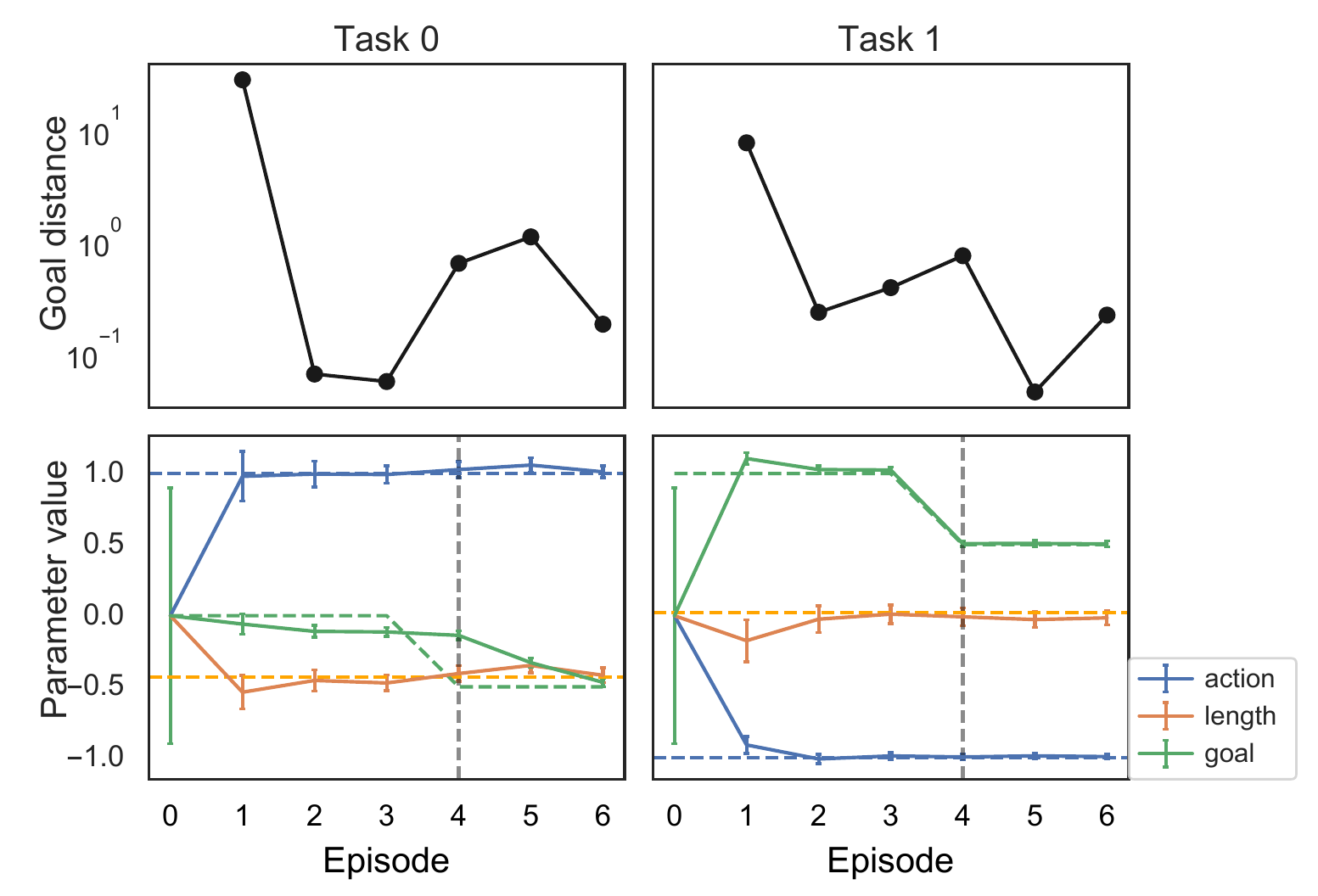}
    \caption{Inference of latent variables when reward hidden parameters are changing but dynamics are not. Left and right columns are two different tasks. \textbf{Top:} Distance to the goal tip position for two CartPole tasks with different hidden parameters (pole length $\in \{0.5, 0.7 \}$.) The first episode is a random policy. \textbf{Bottom:} Mean and standard deviation of the posterior of latent variables (solid), and the true values (dashed). Episode 0 shows the priors on latents before inference.}
    \label{fig:dist_goal}
\end{figure}
To demonstrate inference over a structured latent variable model for a GHP-MDP, we construct a toy example where the goal is to infer the hidden parameter values at test time in order to facilitate accurate planning given the parameterized dynamics function.

For this experiment, we use a modified version of \texttt{CartPoleSwingUp} with a reward function proportional to the distance between the tip of the pole and the desired position $(x_{\mathrm{goal}} - x_{\mathrm{tip}})^2$ \cite{saemundsson2018meta}. The transition function $T(\s, \action; {\eta_a}, {\eta_l}, {\eta_x})$ takes as parameters: $\eta_a$ that scales the action/control signal (blue), the length of the pole ${\eta_l}$ (orange), and the position of the pole tip ${\eta_x}$ (green; Figure~\ref{fig:dist_goal}). We model the tasks by replacing the hidden parameters $\eta$ in the transition model $T_\eta$ with corresponding latent variables ${{\bf z}_a}$, ${{\bf z}_l}$, ${{\bf z}_x}$. In order to use generic priors for missing information, e.g.\ $p(\mathbf{z}) = \mathcal N(0,1)$, we also model unknown positive parameter values $\eta$ with latent variables $\mathbf{z}$ via the softplus: $\eta = \log (1 + \exp(\mathbf{\mathbf{z}}))$.

The experiment consists of only two tasks with different hidden parameters (Figure~\ref{fig:dist_goal} left and right columns). Latent variables are inferred using mean field variational inference with Gaussian priors and variational distributions, and perform control using random search. Because there is no model learning, we can immediately infer latent variables given data. After collecting data for 200 steps using a random policy (episode 1), inference yields accurate estimates (Figure~\ref{fig:dist_goal};~bottom) resulting in good control (Figure~\ref{fig:dist_goal};~top). 

To test the system, the goal position is suddenly changed in episode 4 and see a corresponding change in only the corresponding latent variable when continuing inference. Thus, we demonstrate the ability of a simplified structured latent variable model to properly disentangle variations in an environment at test time through the usage of inference alone and even adapt on the fly to targeted changes by inferring the right component.
}
\subsection{GHP-MDP for continuous control}

We evaluate both the joint and structured LV model with a total of 8 latent dimensions using experiments in the MuJoCo Ant environment, a challenging benchmark for model-based RL \cite{Todorov}.

Because the GHP-MDP models both dynamics and reward, we introduce novel tasks with up to two factors of variation. In \expname{DirectionAnt}, the agent receives reward proportional to the portion of its velocity along the goal direction. This is an example of a multi-task RL setting, often used in meta-reinforcement learning benchmarks, in which tasks $\tau_i$ have common dynamics $\transition$ but a unique reward function $\frew_i$.
In \expname{CrippledLegDirectionAnt}, agents must learn 1)~which of four legs is crippled and does not respond to actions, and 2)~which of eight directions to travel. In both experiments, directions are the eight cardinal plus intercardinal directions. The method for dividing tasks into training, test, and holdout sets is described below.

We experimentally evaluate the joint model from Section~\ref{sec:latent-model} and structured LV model from Section~\ref{sec:plated-model}. The joint model has a global $\lat$ that conditions both dynamics and reward models. The structured model has separate latent variables for dynamics $\tilde{T}(\cdot\,;\latE)$ and reward model $\tilde{\frew}(\cdot\,;\latR)$. Latent variables are 4-D per factor of variation $\lat_i \in \reals^4$ for $i \in \{1, \ldots, K\}$; the joint model has the same total dimensionality in one variable $\lat \in \reals^{4K}$ for K factors of variation. The architecture for all experiments is an ensemble of 5 neural networks with 3 hidden layers of 256 units for the dynamics model, and 1 hidden layer of 32 units for the reward model. We report results averaged across 5 seeds using 95\% bootstrapped confidence intervals.
yielding new combinations with an unseen factors.
\begin{figure}
    \centering
    \includegraphics[width=0.7\columnwidth]{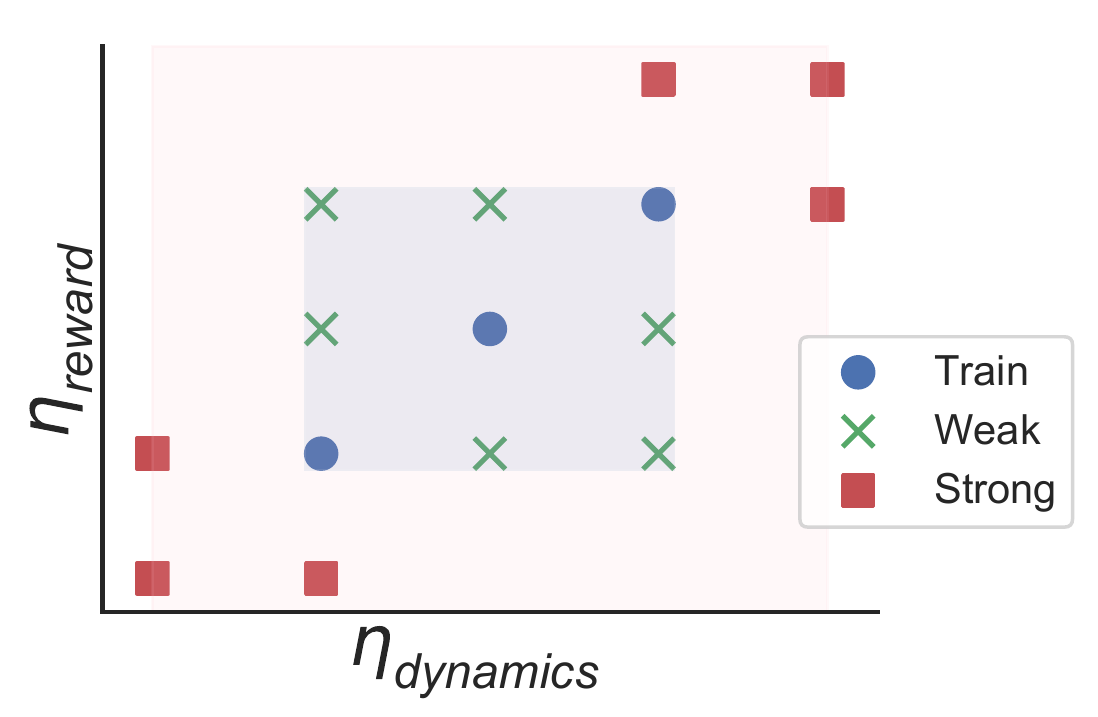}
    \caption{Hidden parameters for training tasks represented by blue dots. Weak generalization requires an agent to work on new combinations (green crosses). Strong generalization requires extrapolation to new hidden parameters (red squares).}
    \label{fig:weak_strong_generalization}
\end{figure}
Experimental results are divided into three categories to probe questions/hypotheses about different flavors of generalization:
\begin{itemize}
    \item \emph{Transfer} that occurs when learning across tasks is faster than learning each task individually. We compare the learning curves of agents trained across tasks to those of specialists trained per task in Section~\ref{sec:transfer}.

    \item \emph{Weak generalization} that requires performing well on a task that was not seen during training but has closely related dynamics and/or reward. Meta-RL commonly assumes tasks at meta-test time are drawn from the same distribution as meta-training, and so falls under this umbrella. In Section~\ref{sec:weak-generalization}, we test on novel \emph{combinations} of crippled leg and goal direction in \expname{CrippledLegDirectionAnt} that were not observed during training.

    \item \emph{Strong generalization} that requires performing well on a task with dynamics and/or reward that is outside what was seen during training. This setting falls under transfer learning or online adaptation, in which an agent leverages previous training to learn more quickly on a new out-of-distribution task/environment. In Section~\ref{sec:strong-generalization}, we test on a novel goal direction not observed during training in \expname{DirectionAnt} and \expname{CrippledLegDirectionAnt}.
\end{itemize}
Within the GHP-MDP formalism, the last two types of generalization can be visualized as in- or out-of-distribution in the space of hidden parameters, as shown in Figure~\ref{fig:weak_strong_generalization}. Another point of interest is how well a structured LV model will generalize compared to the joint LV model. Intuitively, separating the latent variables along causal relationships ought to improve (strong) generalization when samples are scarce.

\begin{figure}[ht]
\centering
    \centering
    \includegraphics[width=.95\columnwidth]{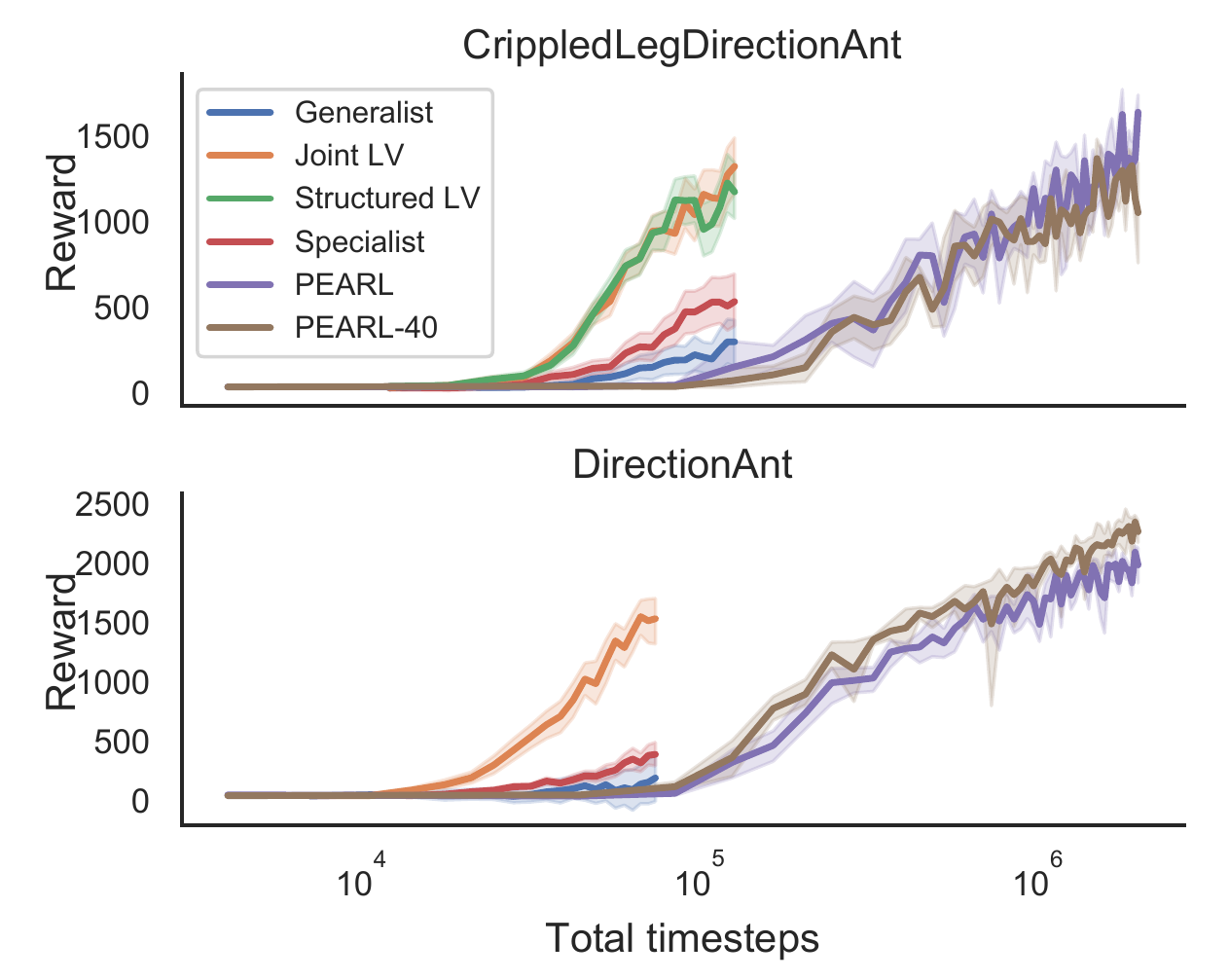}
    \caption{\protect\raggedright Learning curves (semi-log scale) for (bottom) \expname{DirectionAnt} and (top) \expname{CrippledLegDirectionAnt}. "Joint LV" in the legend refers to the only LV model in the bottom panel.}
    \label{fig:reward-training-results}
\end{figure}

We compare our method to three baselines. The {\bf Generalist} is a ensemble-model-based baseline \cite{Chua2018} that learns dynamics and rewards but lacks latent variables. This baseline is a negative control; its failure confirms that the environments are challenging enough to require additional modeling complexity introduced by the GHP-MDP. The \textbf{Specialist} is the same model as the Generalist but trains on each task \emph{individually}, providing a benchmark for per-task sample efficiency. {\bf PEARL} is a sample-efficient off-policy meta-RL algorithm that also uses latent "context" variables and amortized inference \cite{PEARL}, and reports state-of-the-art sample efficiency on continuous control meta-RL tasks like \expname{DirectionAnt}. We note that PEARL can be posed as the model-free analogue of the joint LV model, using an inference network (like a VAE) instead of SVI. (We know of no analogue to the structured LV model.) We allow PEARL $\approx 2M$ samples, up to $\approx 15\times$ the training data as our method. To compare generalization ability when learning from few tasks, {PEARL} learns from the same number of tasks as our model.
Because reward varies across tasks in our experiment, we do not compare against adaptive model-based methods like \cite{Clavera2018} that do not model reward.
Because meta-RL methods like PEARL are usually trained with many more tasks, an additional baseline \textbf{PEARL-40} trains on 40 out of 100 possible tasks. We compare PEARL and PEARL-40 to evaluate the effect of additional tasks, and compare our methods with PEARL trained on the same tasks to evaluate our method's performance.

\subsection{Learning, transfer, and generalization}
\label{sec:transfer}

\begin{figure}[tb]
\centering
\includegraphics[width=.95\columnwidth]{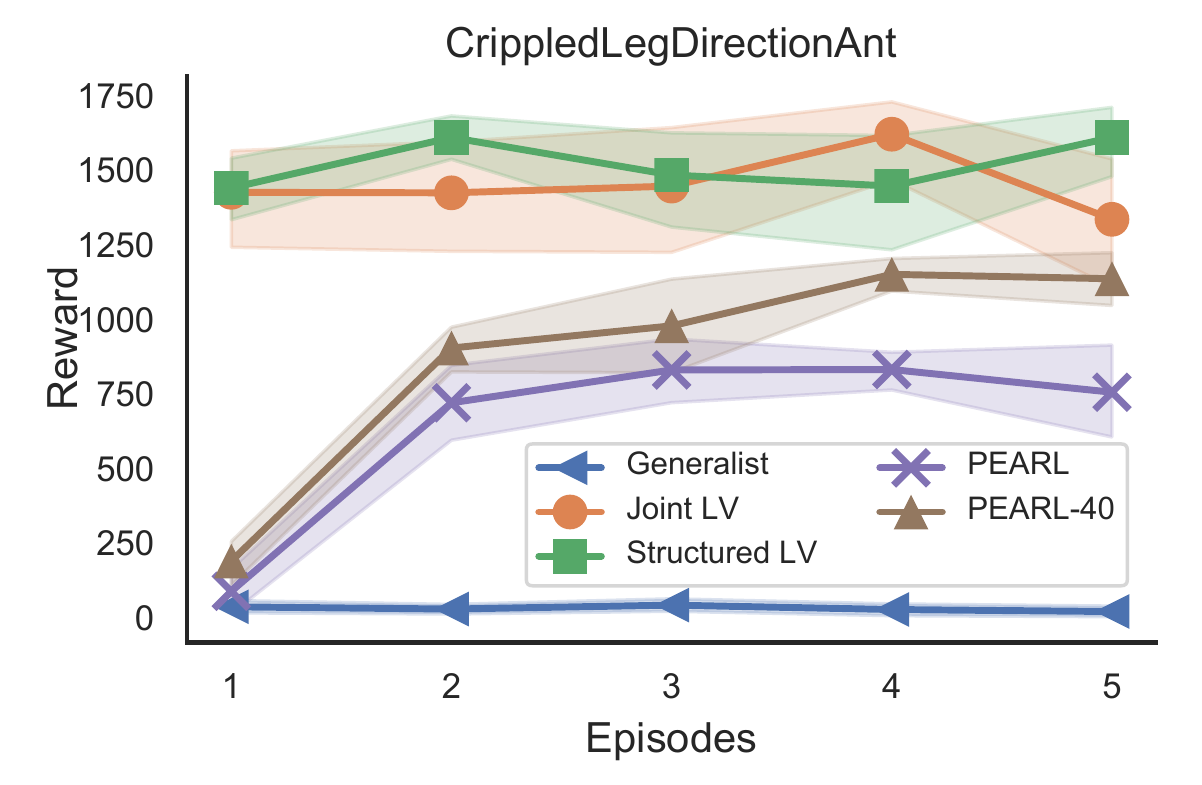}
\caption{Episode reward on novel combinations.}
\label{fig:ant_weak_generalization}
\end{figure}

Of the 28 total tasks (7 directions and 4 crippled legs) in \expname{CrippledLegDirectionAnt}, 12 are sampled for training such that each direction and crippled leg is seen at least once. A sample of 5 remaining combinations is used to evaluate weak generalization in the next section. For \expname{DirectionAnt}, 7 training directions are seen during training.

First, we compare the training performance of the LV models to the Specialist baseline to measure positive transfer. The average performance across tasks is plotted against the total number of timesteps taken across all tasks in Figure~\ref{fig:reward-training-results}. In both experiments, the LV models learn significantly faster across tasks than the Specialist, indicating that the latent variables facilitate information sharing into the global neural network models. This is confirmed by the poor performance of the Generalist that also sees all the tasks but is unable to pool information effectively, yielding agents that perform worse than the Specialist. 
In addition to modeling the reward, our controller also handles early termination of an episode (see Sec.~\ref{sec:control}), whereas other model-based methods fail \cite{wang2019benchmarking}. Accordingly, even the Specialist is state-of-the-art under these conditions, outperforming similar models proposed int he literature on our tasks.
Second, we compare to PEARL benchmarks and observe that the LV models are $>10\times$ more sample efficient than the most efficient off-policy meta-RL algorithm that we are aware of.
Finally, the difference between PEARL and PEARL-40 is small, suggesting that training is unaffected by fewer tasks, but as we will see (Sec.~\ref{sec:weak-generalization}), still negatively impacts generalization.

\subsubsection{Weak generalization:}
\label{sec:weak-generalization}

To test for weak generalization, 5 tasks are sampled from novel combinations of crippled leg and direction in \expname{CrippledLegDirectionAnt} (excluding the holdout direction) and evaluated for 5 episodes at test-time. For LV models, the same objective function (either \eqref{eq:loss-joint} or \eqref{eq:loss-plated}) is minimized except that only variational parameters are updated. Every 10 steps on the test task, 100 iterations of SVI are performed on observed data at $5\times$ the learning rate. 

To address the gap in sample efficiency between model-based and model-free methods, we evaluate all models after the maximum amount of training so that the LV models are compared to PEARL with much more training data. This shifts the focus to the realized reward for a fairer comparison.
Despite far less training data and similar training performance, Figure~\ref{fig:ant_weak_generalization} reveals that both LV models outperform the meta-RL baseline regardless of the training regimen. The LV models also infer quickly, performing well on the first episode, whereas PEARL is designed to perform well after one or two episodes. (Note that we did not attempt to tune PEARL's inference to perform well in the first episode, and ours was tuned to maximize average reward in the first 3 episodes.)
There is, however, no difference between the joint and structured LV models. We hypothesize the latter will scale better with even more factors of variation.
\label{sec:strong-generalization}
\begin{figure}[tb]
    \centering
    \includegraphics[width=.95\columnwidth]{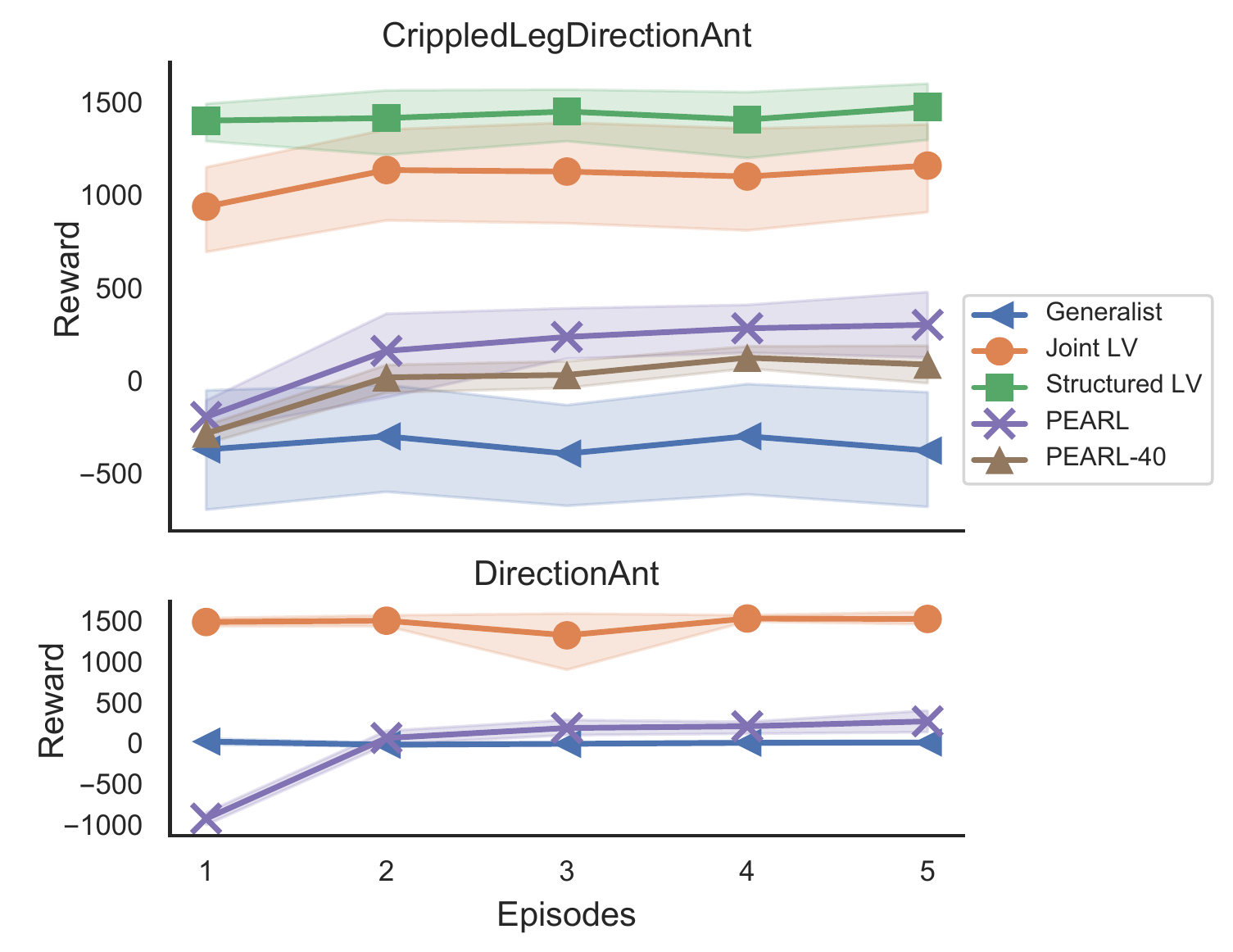}
    \caption{Episode reward on a novel direction.}
    \label{fig:ant_strong_generalization}
\end{figure}

\subsubsection{Strong generalization:}

In this experiment, trained models infer the eight holdout directions at test-time to evaluate generalization to a novel task. In \expname{CrippledLegDirectionAnt}, the agent must also infer which leg is crippled, further raising the difficulty (Recall that each leg was crippled at least once during training).

Figure~\ref{fig:ant_strong_generalization} shows the LV models significantly outperform the baselines, supporting the claim that the GHP-MDP effectively models variations across these tasks. The structured LV model is the best performing, consistent with our hypothesis that causally factorized latent variables can improve generalization under some circumstances. Surprisingly, PEARL trained with fewer tasks fares slightly better than PEARL-40, indicating that a model trained with more tasks, which aids weak generalization to similar tasks (Figure~\ref{fig:ant_weak_generalization}, slightly biases the policy against out-of-distribution tasks that require extrapolation.

\section{Discussion}

In this work we have introduced the GHP-MDP, which can capture hidden structure in the dynamics and reward functions of related MDPs.
We demonstrate this modeling approach on continuous control tasks with dynamics and reward variations that surpass strong baselines in performance and sample efficiency. 
In future work, it would be interesting to study the extent to which one can model more factors of variation and disentangle them automatically, obviating the specification of structure upfront.

% References and End of Paper
% These lines must be placed at the end of your paper
\small
\bibliography{pmbrl}
\bibliographystyle{aaai}
\end{document}